\documentclass{article}

\usepackage[utf8]{inputenc} 
\usepackage[T1]{fontenc}    
\usepackage{hyperref}       
\usepackage{url}            
\usepackage{booktabs}       
\usepackage{amsfonts}       
\usepackage{nicefrac}       
\usepackage{microtype}      
\usepackage{lipsum}
\usepackage{fancyhdr}       
\usepackage{graphicx}       
\graphicspath{{media/}}     

\usepackage[margin=1in]{geometry}

\usepackage{bm}

\usepackage{enumitem}
\usepackage{multirow}

\usepackage[ruled,vlined]{algorithm2e} 
\SetKwInOut{Input}{input} 
\SetKwInOut{Output}{output}  

\usepackage{makecell}

\usepackage{amsmath,amssymb}
\usepackage{cleveref}

\usepackage{xcolor}
\newcommand{\edit}[1]{\textcolor{black}{#1}}

\title{\textit{Sim2Real} in Reconstructive Spectroscopy: \\ 
Deep Learning with Augmented Device-Informed Data Simulation
}

\author{
  Jiyi Chen*,\quad Pengyu Li*,\thanks{The first two authors contributed equally to the work.} \quad Yutong Wang,\quad Pei-Cheng Ku,\quad Qing Qu \\
  Department of Electrical Engineering \& Computer Science \\
  University of Michigan, Ann Arbor \\
  \texttt{\{jiyichen, lipengyu, yutongw, peicheng, qingqu\}@umich.edu} \\
}

\begin{document}
\maketitle

\begin{abstract}
This work proposes a deep learning (DL)-based framework, namely \textit{Sim2Real}, for spectral signal reconstruction in reconstructive spectroscopy, focusing on efficient data sampling and fast inference time. The work focuses on the challenge of reconstructing real-world spectral signals under the extreme setting where only device-informed simulated data are available for training. Such device-informed simulated data are much easier to collect than real-world data but exhibit large distribution shifts from their real-world counterparts. To leverage such simulated data effectively, a hierarchical data augmentation strategy is introduced to mitigate the adverse effects of this domain shift, and a corresponding neural network for the spectral signal reconstruction with our augmented data is designed. Experiments using a real dataset measured from our spectrometer device demonstrate that \textit{Sim2Real} achieves significant speed-up during the inference while attaining on-par performance with the state-of-the-art optimization-based methods.
\end{abstract}


\tableofcontents

\section{Introduction}\label{section:Intro}
Optical spectroscopy is a versatile technique for various scientific, industrial, and consumer applications. Recently, computational spectroscopy using reconstructive algorithms \cite{chang2008estimation,Kurokawa2011,bao2015colloidal, Wang2014e, Wang2019,rs2, Sarwar2020,UCLA_paper} has been rapidly developed owing to its potential to enable a miniaturized spectrometer \cite{chip_rs1, chip_rs2}. A spectrometer encodes optical spectral information spatially or temporally and then measures the encoded information using a series of photodetectors. The transformation between the input signal $\bm{x}$ and the photo-detector readout $\bm{y}$ in a spectrometer design can be either linear or nonlinear. However, a linear design is typically preferred. An example of a nonlinear encoding is observed in Fourier-Transform infrared (FTIR) spectrometers, which produce an auto-correlation of the input through a variable time-delay interferometer. For a linear system, $\bm{x}$ and $\bm{y}$ can be represented using column vectors while the mapping is represented by a responsivity matrix $\bm{R}$ such that $\bm{y}=\bm{Rx}$. For example, a monochromator spatially disperses different spectral components with a block-diagonal $\bm{R}$ matrix. In a reconstructive spectrometer, $\bm{R}$ is generally much more complex. Each photo-detector's readout depends on all instead of a few spectral components. While the complex structure of $\bm{R}$ means a lot of computational resources are generally required to recover the spectrum $\bm{x}$ from the photo-detector readouts $\bm{y}$, it also opens up new opportunities. These include the ability to tailor $\bm{R}$ to encode the spectral information directly to parameters of interest in the application, such as the spectral peak positions and the relative intensities between the peaks; the ability to increase the accuracy of spectral reconstruction by focusing on the most important spectral information even when $\bm{y}$'s dimension is much smaller than $\bm{x}$'s\cite{candes2006stable}; and the ability to miniaturize the spectrometer. 

A wealth of research has been conducted to increase spectral reconstruction's efficiency and accuracy. The approaches can generally be categorized as optimization-based or data-driven.
The optimization-based approaches formulate the inverse problem into a convex optimization problem in terms of the signal to be reconstructed. To deal with non-uniqueness of the solution, different types of regularizations (e.g., nonnegativity, sparsity, and smoothness) have been considered \cite{RUDIN1992259, NNLS_TV,chang2008estimation}.
The effectiveness of these optimization-based techniques often hinges on the precise adjustment of regularization parameters and hyperparameters, a task that can be challenging \cite{Li23}. Moreover, optimization-based approaches based on iterative procedures can be energy and computationally expensive for critical applications such as Internet-of-Things or time-sensitive tasks. Despite the challenges, miniaturized spectrometers based on these spectral-reconstruction techniques have reported good performance. For example, we recently developed a chip-scale spectrometer concept by integrating 16 semiconductor photo-detectors, each with a different spectral response \cite{NNLS_TV}. We designed the photo-detectors using nanostructured materials to minimize the dependence on the incident angle of light and thus eliminated the need for any external optical components. The entire spectrometer, sans the electronic readout circuitry, is only a few microns thick. As the photodetector's response was broadband, the device was able to recover basic spectral information, including the positions and relative intensities of peaks across the entire visible spectrum. The accuracy of reconstructing a multi-peak spectrum were 0.97\% (RMS error) for locating the peak locations, comparable to a monochromator with twice the number of detectors. Enabled by the rich interaction between different spectral components in each photo-detector, we also showed that only seven detectors were sufficient for input signals with up to three peaks and had relatively smooth spectra.

\begin{figure}[t]
\centering 
\includegraphics[width=\linewidth]{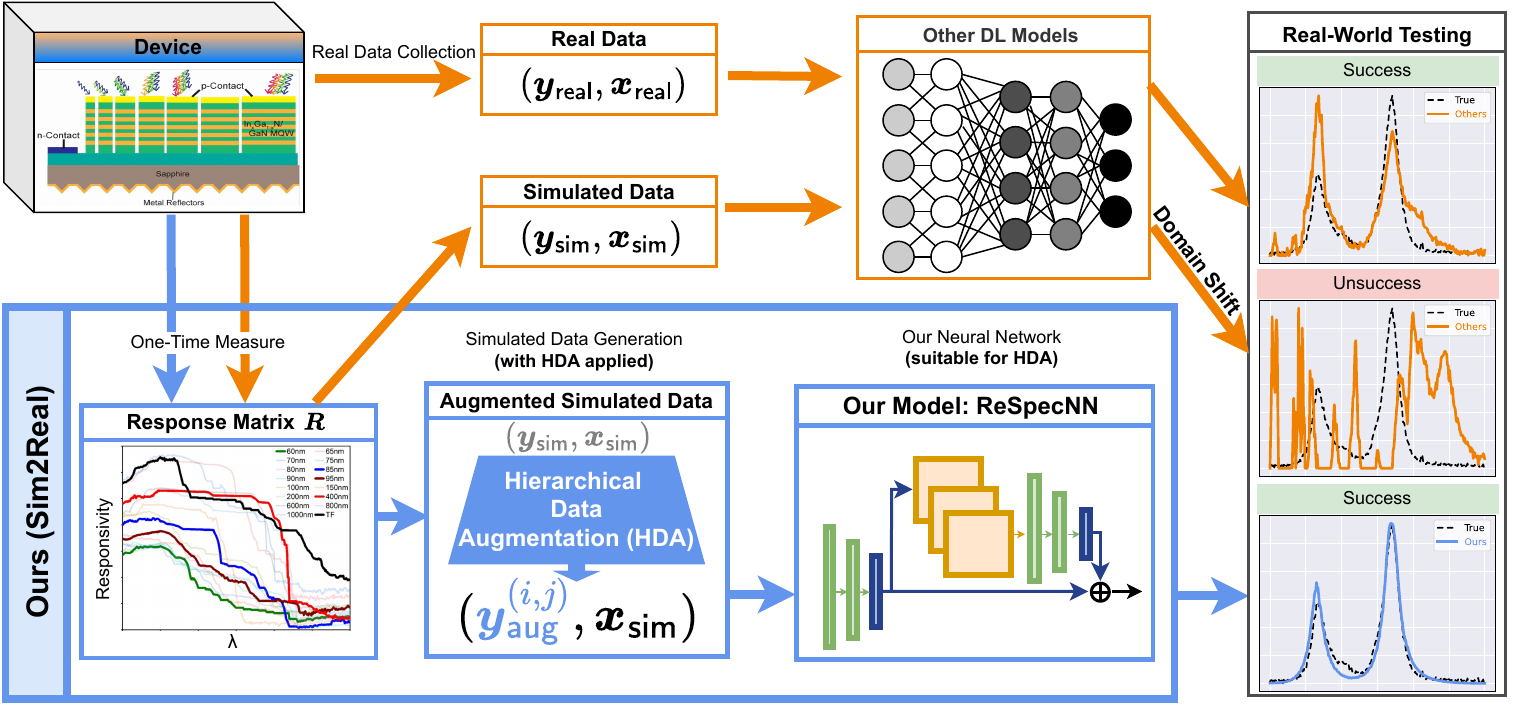}
\caption{\textbf{The diagram of our proof-of-concept \textit{Sim2Real} framework in the reconstructive spectroscopy.}
The orange arrows denote existing methods, which require collecting and training on real-world data. The blue box and arrows denote our proposed \textit{Sim2Real} framework that effectively address the domain shift between simulated and real-world data. \edit{The domain shift is visualized through PCA in \Cref{fig:pca}} See Section~\ref{section:methods} for details. \edit{The Response Matrix plot is reprinted with permission from [13]. Copyright 2022 American Chemical Society.} 
}
\label{fig:intro}
\end{figure} 

In comparison, deep learning-based data-driven approaches \cite{UCLA_paper, kim2022compressive, zhang2020denoising} showed the potential to address the challenges of an optimization-based approach. First, deep overparametrized networks trained by gradient descent methods tend to enforce the regularization implicitly\cite{arora2019implicit}, avoiding the need to adjust regularization parameters. Second, while training neural networks can be computationally expensive, reconstructing a new measurement during inference requires only a single forward pass through the network. In contrast, optimization-based approaches can be costly in terms of energy and computing resources during inference. Various data-driven methods of reconstructive spectroscopy have been developed, including utilizing a simple, fully connected network on a plasmonic spectral encoder on the device\cite{UCLA_paper}, minimizing potential noise before the reconstruction process on a colloidal quantum dot spectrometer\cite{zhang2020denoising}, and a specifically tailored residual convolutional neural network aiming to improve the reconstruction performance\cite{kim2022compressive}.

A significant challenge with the current deep learning methodologies is their dependence on extensive volumes of real, precisely annotated training data to achieve optimal performance\cite{UCLA_paper, Li23}.  In a practical setting, gathering such real, labeled data for spectrometer reconstruction is both costly and time-intensive. Additionally, the actual labeled data pairs collected often contain considerable noise, which may result in poor performance during testing when used to train deep learning models.

In this work, we introduce a method to tackle these challenges in deep learning-based approaches for spectral reconstruction. As illustrated in \Cref{fig:intro}, we propose a \textit{Sim2Real} framework in which we train the deep learning models \emph{solely} based on simulated datasets and then deploy the model on a real dataset. Our method contains two key components:
\begin{itemize}[leftmargin=*]
    \item \textbf{\emph{Hierarchical Data Augmentation (HDA):}} To mitigate the domain gap between simulated and real data, we introduce a data augmentation technique to perturb both the response matrix $\bm{R}$ and the encoded signal. Training the model with these data improves its robustness. 
    \item \textbf{\emph{ReSpecNN:}} In line with this, we developed a new lightweight network architecture specifically designed for the \textit{Sim2Real} framework, which is applied to the spectral reconstruction problem with the aforementioned HDA approach. 
\end{itemize}

Unlike conventional methods, marked by orange arrows in \Cref{fig:intro}, which require collecting and training on real-world data, our strategy eliminates the need for extensive real-world data collection, only requiring a single measurement of the response matrix $\bm{R}$. Moreover, by utilizing our augmented simulated training data, we effectively close the domain gap between simulated and real data, leading to high-quality spectral reconstruction during testing on real data.

The rest of the paper is organized as follows. In \Cref{section:methods}, we introduce the mathematical formulation of the spectral reconstruction problem while identifying the limitations of an existing method. In \Cref{sec:our-contributions}, we present our main contributions and discuss how they address the limitations of the existing method. In \Cref{section:experiments}, we validate the performance of our proposed method on real-world data by comparing it with the state-of-the-art method and discussing the implications.

\section{Problem formulation and existing Methods}\label{section:methods}

In this section, we introduce the mathematical formulations of the spectrum reconstruction problem. We also identify the challenges and limitations of existing optimization and deep learning method.

\subsection{Spectral Reconstruction Problem}

For spectrometer signal reconstruction, the \emph{encoded signal vector} $\bm{y} = (y_1,\dots, y_K) \in \mathbb{R}^K_{+}$ is the output produced by the spectrometer given the signal of interest \(\bm{x}\) as input, and is modeled mathematically as: 
\begin{align}
    y_i &\textstyle= \int_a^b \bm{x}(\lambda) R_i(\lambda)\,d\lambda + \epsilon_i  
    \\&\textstyle
    \approx \sum_{j=1}^{\ell} \bm{x}(\lambda_j) R_i(\lambda_j) \Delta\lambda + \epsilon_i, \, \,\, \mbox{ for } i=1, \ldots, K, \label{eq:linear_system}
\end{align} 
where \Cref{eq:linear_system} 
is the discrete approximation of the model. Here, we denote the \emph{spectral signal of interest} at the wavelength \(\lambda\) by $\bm{x}(\lambda) \in \mathbb{R}_{+}$, where \(\lambda\) belongs to some predetermined wavelength range of interest $[a, b]$. The signal is encoded, i.e., measured, by \(K\) distinct \emph{spectral encoders} with various \emph{responsivity} at different wavelengths as shown in \Cref{fig:intro}. We denote $R_i(\lambda)$ as the \emph{relative spectral power density} of the $i$-th spectral encoder and $\epsilon_i$, the error of the $i$-th encoder, where \(i \in \{1,\dots, K\}\). 
The error term, $\epsilon_i$, is a random quantity whose dependency on \(\bm{x}\) is not assumed to be known.

As shown in \Cref{eq:linear_system}, sampling wavelengths $\{\lambda_1,\dots, \lambda_\ell\}$ results in the equally spaced discretization of \([a,b]\)  with a spectral resolution $(b-a)/\ell$. 
Stacking up the relative spectral power density for each encoder gives us the response matrix
$
\bm{R} := \begin{bmatrix}
    R_{ij}
\end{bmatrix}_{i= 1, \, j=1}^{K, \,\ell} \in \mathbb{R}^{K \times \ell}
$
where
\(
\bm{R}_{ij} := R_i(\lambda_j)
\). As such, the discretization relationship between $\bm{y}$ and $\bm{x}$ could be expressed through $\bm{R}$ in a cleaner form:
\begin{align} \label{eq:problem_statement}
    \bm{y} = \bm{R} \bm{x} + \bm{\epsilon}, \quad 
    \mbox{ where } \bm{\epsilon} = [\epsilon_1, \cdots, \epsilon_K]^\top 
\end{align}
As such, our goal is to recover \(\bm{x}\) from observed \(\bm{y}\) and \(\bm{R}\) under the setting where the number of encoders $K$ is much smaller than the number of sampled wavelengths $\ell$. In other words, the system in \Cref{eq:problem_statement} is highly under-determined with non-unique solutions.

\subsection{Optimization-Based Approaches and Limitations}

Note that given the response matrix $\bm R$, the reconstruction problem in (\ref{eq:problem_statement}) can be approached using (non-negative) least squares, as discussed in the introduction. Due to the ill-posedness of the problem,  \emph{explicit} regularization techniques are used to ensure unique solutions with desirable additional structures. Within optimization-based methods, the least-squares (LS) estimate $\hat{\bm{x}} \in \mathrm{argmin}_{\bm{x}} \|\bm{y} - \bm{Rx} \|_2^2$ fails to consider the signal's nonnegativity, often leading to suboptimal solutions. This issue can be addressed by employing nonnegative least squares (NNLS) \cite{chang2008estimation}. To deal with an underdetermined system, regularization strategies like $\ell_1$-norm, $\ell_2$-norm, or total-variation (TV) norm regularization have also been proven effective\cite{RUDIN1992259, NNLS_TV}.

However, solving these optimization problems can be time and memory consuming and their performance is sensitive to the choice of the regularization hyperparameter, so that they are often too expensive to deploy on chip-scaled devices. 

\subsection{Deep Learning Methods and Challenges}\label{section:data_gen}


Recently, deep learning-based approaches gained popularity due to their modeling flexibility and fast inference speed. However, a major bottleneck for effectively leveraging deep neural network models is gathering a large amount of real spectral data pairs $(\bm{y}_{\textsf{real}}, \bm{x}_{\textsf{real}})$, a task that is both expensive and time-consuming in reconstructive spectroscopy. 

\begin{figure}[t]
\centering 
\includegraphics[width=0.5\linewidth]{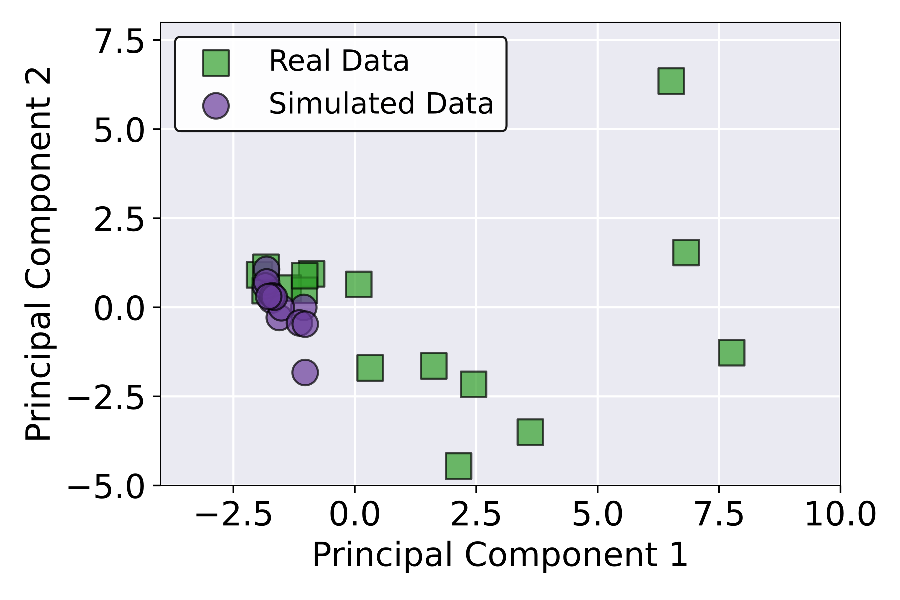}
\caption{\textbf{The PCA Projection of Simulated and Real Data}. The clear separation between clusters along the first two principal components highlights the distribution differences between the simulated data $\bm{y}_{\textsf{sim}}$ and real data $\bm{y}_{\textsf{real}}$, indicating a significant domain shift.}
\label{fig:pca}
\end{figure}

Therefore, based on the problem formulation in \Cref{eq:problem_statement}, previous approaches have focused on using the response matrix $\bm{R}$ to produce simulated spectral data pairs $(\bm{y}_{\textsf{sim}}, \bm{x}_{\textsf{sim}})$, aiming to mimic the distribution of real data obtained in laboratory settings. 
\edit{However, we observe that the simulated data and the real-measured data exhibit an undesirable phenomenon known as ``domain shift'', leading to the poor performance of model trained solely on simulated data when applied to real ones. To provide a straightforward quantitative assessment of the domain shift in our problem, here we ran PCA on both simulated data and real data (preprocessed by \emph{log-min-max} as mentioned in Section \ref{section:experiments}) to visualize the discrepancy via their two principal components. As depicted in \Cref{fig:pca}, the real data presents a much larger spread and variation along both axes compared to the simulated data, which demonstrated the existence of the domain shift.}

Later in \Cref{sec:our-contributions}, we delve into this issue and detail our approach to bridging the domain shift. For reader's convenience,  we first review previous approaches in detail.

\paragraph{Training on Simulated Data.}
In practice, the simulated spectral signals $\bm{x}_{\textsf{sim}}$ can be viewed as the sum of Lorentzian distribution functions \cite{kim2022compressive, NNLS_TV}.
Given the response matrix $\bm{R}$, a simulated spectral signal $\bm{x}_{\textsf{sim}}$ and its corresponding encoded signal $\bm{y}_{\textsf{sim}}$ can be generated as follows:

\begin{enumerate}
    \item Generate $M$ single peak Lorentzian distribution functions independently with various standard deviations.
    \item Sum and then normalize the heights of those Lorentzian curves within range $[0, 1]$. We denote the result distribution as $\bm{x}_{\textsf{sim}}$. 
    \item Multiply $\bm{x}_{\textsf{sim}}$ with the response matrix $\bm{R}$ to produce the encoded signal $\bm{y}_{\textsf{sim}} = \bm{R} \bm{x}_{\textsf{sim}}$. 
\end{enumerate}

Specifically, in the first step, each (Lorentzian) peak  is characterized by three parameters: the mean $\mu$, the width constant $\gamma$, and the intensity constant $I$. The parameters correspond to the peak location, spectral width, and intensity magnitude, respectively. Each parameter is sampled i.i.d uniformly from a set of ranges to be chosen to match the specific characteristics of the spectrometer device. Under our problem setting, the ranges are set to $\mu \in [0,205]$, $\gamma \in [15,20]$ and $I \in [0.25,1]$ respectively.

\paragraph{Challenges in Deploying Trained Models on Real Data.} Equation \eqref{eq:problem_statement} is a simplified model of the actual spectroscopy system. 
As such, the procedure in the previous section for generating simulated spectral signals $\bm{x}_{\textsf{sim}}$ and their corresponding encoded signals $\bm{y}_{\textsf{sim}}$ may produce distributions that differ from the distribution of real encoded signals $\bm{y}_{\textsf{real}}$. This may be attributed to the circuit design, which, in reality, may introduce diverse types of unknown noise into the system, resulting in this distributional difference between real and simulated data. When applying machine learning algorithms, this difference, or ``domain shift'', could cause degraded performance. 

\begin{figure}[htbp]
\centering 
\includegraphics[width=0.5\linewidth]{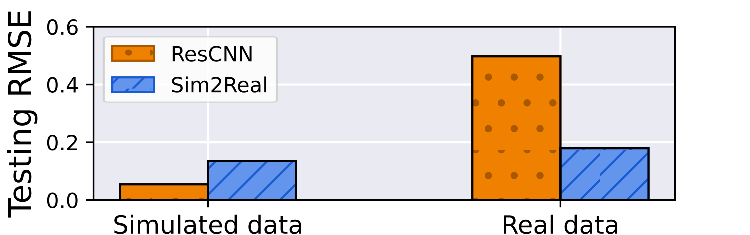}
    
\caption{\textbf{The testing RMSE on the simulated and real data}\cite{NNLS_TV} for \textit{ResCNN}~\cite{kim2022compressive} and our proposed model \textit{ReSpecNN}. Both models followed the \textit{Sim2Real} training setting, that is, trained solely on simulated data. Our model further incorporated the hierarchical data augmentation during training, while \textit{ResCNN} does not.}
\label{fig:comparison-with-kim2022compressive}
\end{figure}

We illustrate this in \Cref{fig:comparison-with-kim2022compressive}, where we train a model exclusively on simulated data that can successfully reconstruct the spectrum given simulated data as input. However, the model's performance degrades significantly on real-world data, even when the real-world and simulated data appear visually similar. Detailed empirical evidences are discussed in Section~\ref{subsec:diss_domain_gap}.

\subsection{Domain Shift between ``Sim'' and  ```Real''}\label{sec:domain-gap}

The challenges of ``domain shift'' are common for deep learning-based approaches.
The ``domain shift'' or ``domain gap'' issue was first observed in transfer learning\cite{pan2009survey}, where a model trained on a large source dataset is fine-tuned on a smaller, more specific dataset.  
To mitigate the performance degradation due to distributional discrepancy between source and target datasets, many techniques have been proposed, including domain adaptation\cite{tzeng2017adversarial,ganin2015unsupervised,zhang2018collaborative}, meta-learning\cite{finn2017model}, and few-shot learning\cite{vinyals2016matching, snell2017prototypical}. These techniques have been shown to be effective in reducing the domain gap in various fields including computer vision\cite{girshick2014rich, zoph2018learning} and natural language processing\cite{dai2015semi, howard2018universal, devlin2018bert}. 

However, when applied to the spectroscopy reconstruction, those proposed methods fall short of solving the domain gap for the following reasons. First, we do not assume access to \emph{any} data from the target distribution. Existing methods still require a certain amount of data from the targeted domain for fine-tuning, which is not applicable under our assumption. \edit{Second, most domain adaptation methods have been designed for classification-based tasks, making them less directly applicable to our spectral reconstruction problem in spectroscopy, which is fundamentally a (non-negative) regression problem. While a few existing results focus on practical regression problems \cite{chen2021representation, nejjar2023dare}, they often deal with simple regression problems rather than tackling the more challenging inverse problem setting which we consider.}
\edit{Third, regarding the topic of deep learning for inverse problems, for tasks like \emph{image} reconstruction, existing research primarily leverages the inductive bias of  neural network architectures suitable for \emph{image} signals \cite{dong2015image, kulkarni2016reconnet, ongie2020deep}. In this work on \emph{spectral} reconstruction, we deal with one-dimensional \emph{spectral} signals which requires a different model design.} 

\edit{While deep learning approaches have become commonly used for spectral reconstruction problems, few studies tackle such domain shift issues directly. A recent work\cite{UCLA_paper} successfully reconstructed spectra with up to 14 peaks using a model trained on spectra with up to 8 peaks under the blind testing conditions, the training and test data are still drawn from similar conditions and distributions.}
Therefore, a specialized method is required to tackle the domain shift issue in solving the spectral reconstruction problem. We now introduce the proposed \textit{Sim2Real} framework in the following.

\section{Our \emph{S\MakeLowercase{im}2R\MakeLowercase{eal}} Framework}\label{sec:our-contributions}

In this section, we introduce the following \textit{Sim2Real} framework to bridge the domain shift between the simulated encoded signal data and the data from real-world spectrometers. Our method tackles the domain shift by two key components: (\emph{i}) hierarchical data augmentation for training data generation, and (\emph{ii}) a lightweight network architecture designed for the spectrum reconstruction problem with our HDA.

\vspace{-0.1in}
\subsection{Hierarchical Data Augmentation}

Although we measure the response matrix $\bm{R}$ in advance and consider it to be fixed and known, our hierarchical data augmentation strategy acknowledges the potential uncertainty in our measurement of $\bm{R}$ and the encoded signal vector $\bm{y}$ to improve model robustness and minimize the domain gap. For every pair of simulated training data $(\bm{y}_\textsf{sim}, \bm{x}_\textsf{sim})$ introduced in \Cref{section:data_gen}, we systematically introduce noise as outlined in \Cref{algorithm:data_aug}.

\begin{figure}[ht]
\centering 
\includegraphics[width=\textwidth]{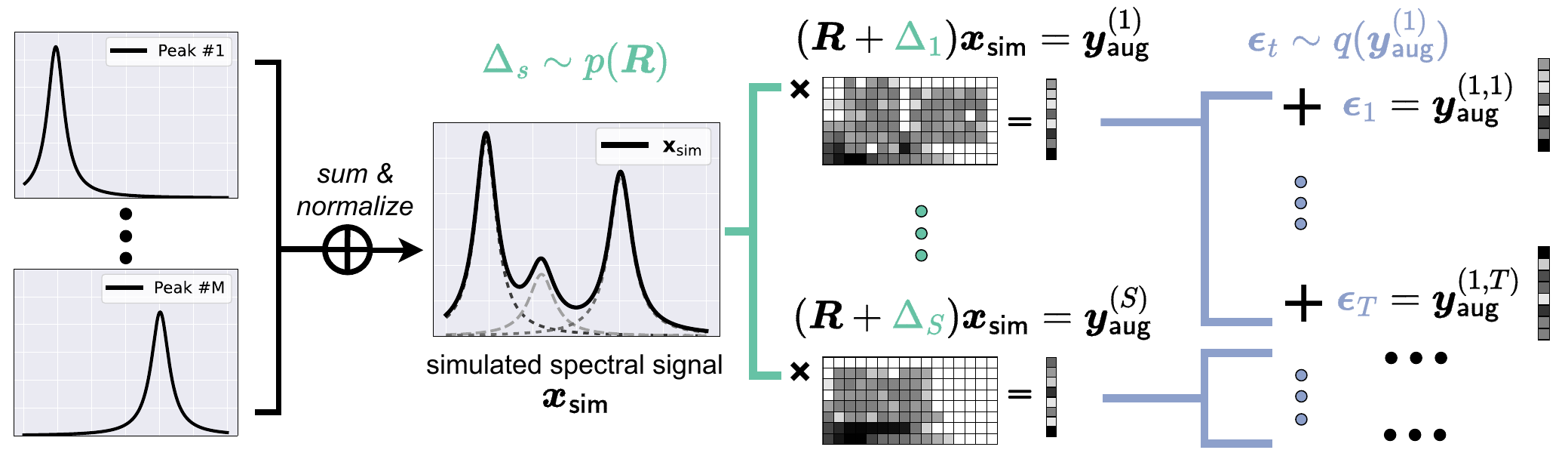}
\caption{\textbf{The diagram of the simulated data generation procedure and our proposed hierarchical data augmentation (HDA) scheme.} 
Simulated spectral signal $\bm{x}_{\textsf{sim}}$ is generated through the sum of Lorentzian distribution. For one given $\bm{x}_{\textsf{sim}}$, we generate many corresponding augmented encoded signals $\bm{y}^{(S, T)}_{\textsf{aug}}$ by adding noise $\Delta_{S}$ to $\bm R$ before multiplying with the spectral signal and adding noise $\bm{\epsilon}_T$ afterward. The two noise distributions could be chosen flexibly. This HDA process is summarized in detail by \Cref{algorithm:data_aug}}
\label{fig:data_generation}
\end{figure}

To illustrate this process, we have provided a visualization in \Cref{fig:data_generation}, using an example of 3LED samples (number of peaks $M=3$) explained in detail below.
Instead of simply multiplying response matrix $\bm{R}$, our hierarchical data augmentation extends each original simulated data sample $(\bm{y}_{\textsf{sim}}, \bm{x}_{\textsf{sim}})$ to $S\times T$ augmented ones $(\bm{y}_{\textsf{aug}}^{(s,t)}, \bm{x}_{\textsf{sim}})$ by adding different noises on the measured response matrix $\bm{R}$ (aquamarine green traces) and each intermediate augmented encoded signal (illustrated by $\bm{y}_{\textsf{aug}}^{(1)}$ only, light purple traces) respectively.


\begin{figure}[ht]
\centering
\begin{minipage}{.6\linewidth}
    \begin{algorithm}[H]
\caption{Hierarchical Data Augmentation (HDA)}
\label{algorithm:data_aug}

\SetAlgoLined 
\DontPrintSemicolon 
\KwIn{\\
\hspace{1em} $\bm{x}_\textsf{sim}$: a simulated spectral signal from \Cref{section:data_gen} \\
\hspace{1em} $\bm{R}$: the response matrix \\
\hspace{1em} $p(\bm{R})$: the distribution of noise perturbation on $\bm{R}$ \\
\hspace{1em} $q(\bm{y^{(s)}_{\textsf{aug}}})$: the distribution of noise perturbation on $\bm{y^{(s)}_{\textsf{aug}}}$ \\
\hspace{1em} $S, T$: the number of perturbations \\
}

\KwOut{\\
\hspace{1em} $(\bm{y}_{\textsf{aug}}^{(1, 1)}, \bm{x}_\textsf{sim}), \cdots, (\bm{y}^{(S, T)}_{\textsf{aug}}, \bm{x}_\textsf{sim})$: training data pairs. \\
}
\For{$s=1,\cdots,S$}{
    Sample $\bm{\Delta}_s \sim p(\bm{R})$ \; 
    $\bm{R}_s = \bm{R} + \bm{\Delta}_s$ \;
    $\bm{y}^{(s)}_\textsf{aug} = \bm{R}_s \bm{x}_\textsf{sim}$ \;
    \For{$t=1,\cdots,T$}{
        Sample $\bm{\epsilon}_t \sim q(\bm{y^{(s)}_{\textsf{aug}}})$ \;
        $\bm{y}^{(s, t)}_\textsf{aug} = \bm{y}^{(s)}_\textsf{aug} + \bm{\epsilon}_t$ \;
    }
}
\end{algorithm}
\end{minipage}
\end{figure}

To perturb the response matrix $\bm R$, we simply add Gaussian noise with zero mean and variance related to entries of $\bm{R}$. That is, for each noisy perturbation matrix $\bm \Delta_s$ ($s=1,\cdots,S$),  its $(i, j)^{th}$ entry is sampled i.i.d from the distribution $\mathcal{N}\left(0,\ \sigma^2_{ij}\right)$ with $\sigma_{ij} = \alpha \cdot R_{ij}$, where $\alpha$ is a hyperparameter controlling the intensity of perturbation on $\bm{R}$.  
For perturbing the encoded signal $\bm y$, we inject nonnegative noise $\bm{\epsilon}_t$ ($t=1,\cdots,T$). Here, each entry of $\bm{\epsilon}_t$ is sampled i.i.d from the Gaussian distribution and then passed through the $\textsf{ReLU}(z) = \max\{0,z\}$ operator to enforce non-negativity. \edit{In practice, we determine those hyper-parameters through the empirical experiments. Details could be found in the later \cref{subsec:exp_setup}.}

In the case above, we considered adding Gaussian noise to disrupt the data, which is simple yet effective in practice. However, should specific information about the noise be accessible under certain conditions, we can further refine both distributions $p(\bm{R})$ and $q(\bm{y}^{(s)}_{\textsf{aug}})$, for the noise sampled on the response matrix and the intermediate augmented encoded signal, respectively. 
As a result, for each ground truth spectrum input $\bm{x}_{\textsf{sim}}$, $S \times T$ many corresponding augmented encoded signal vector $\bm{y}_{\textsf{aug}}^{(\cdot,\cdot)}$ are generated. This also demonstrates the generalizability and flexibility of our data augmentation method.
By incorporating structured noise into the device-informed simulated data generation, we term this process \emph{Hierarchical Data Augmentation} (HDA).

\subsection{Network Architecture: ReSpecNN}

\begin{figure}[t]
    \centering
    \includegraphics[width=\textwidth]{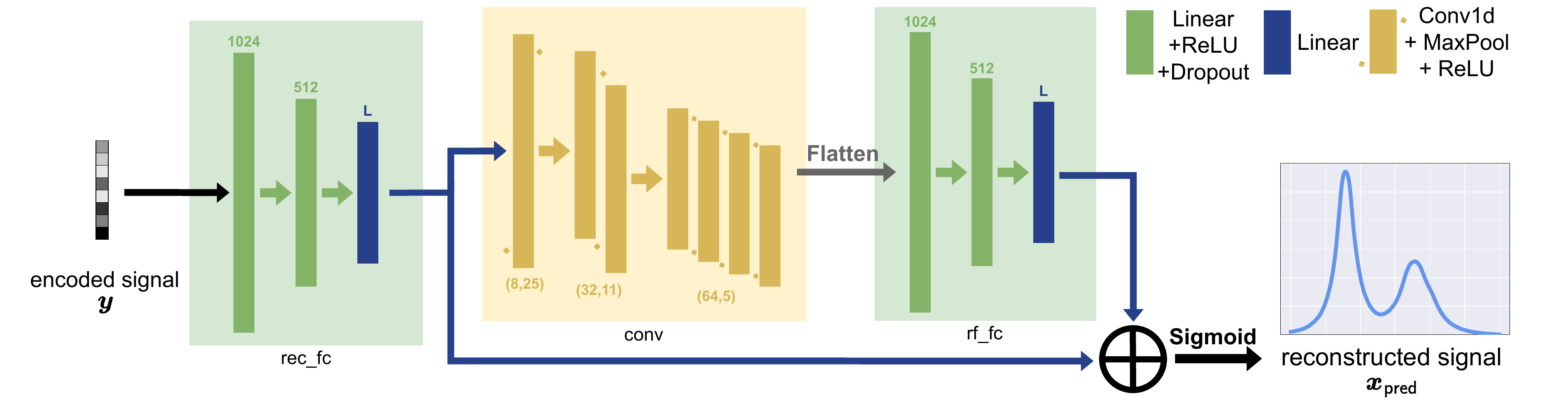}
    
    \caption{\textbf{The architecture of our proposed neural network}. \textit{ReSpecNN} comprises two fully-connected modules (dubbed as \texttt{rec\_fc} and \texttt{rf\_fc}) and a three-layer convolutional neural network (dubbed as \texttt{conv}). Note that \texttt{rec\_fc} and \texttt{rf\_fc} have a residual connection. For each linear layer in fully-connected modules, the number above represents its output dimension, where $L$ denotes the number of input wavelengths. For each 1D convolutional layer, the tuple below specifies the number of filters and the kernel size.}
    \label{fig:NN-architecture}
\end{figure}

For training with augmented data generated by HDA, we propose a deep neural network architecture tailored for spectrometer signal reconstruction, hereinafter referred to as \textit{ReSpecNN}. The architecture is visualized in \Cref{fig:NN-architecture} and explained in detail below.

\edit{The \textit{ReSpecNN} model is specifically tailored for our HDA scheme. In contrast to the previous model by Kim et al. \cite{kim2022compressive}, we incorporate an additional fully-connected block with skip connections to enhance the adaptability of our model in handling HDA-perturbed data. Instead of directly absorbing the information from the measured response matrix $\bm{R}$ into the calculations, which could be misleading due to inaccuracies in $\bm{R}$ measurements, we aim for the extra fully-connected block to potentially learn a more robust inverse of $\bm{R}$ through our HDA-perturbed data. This, in turn, enhances the overall model's robustness, leading to improved reconstruction performance. Specifically, it's important to note that the previous model by Kim et al. \cite{kim2022compressive} initialize input data with information directly from measured $\bm{R}$ which might contain noise or misleading information. In comparison, through HDA, our \textit{ReSpecNN} autonomously learns a more robust $\bm{R}$ for training.}

Our model consists of two fully-connected modules and a 1D convolutional module. The first fully-connected module (\texttt{rec\_fc}) aims to construct each spectrum at a gross level. To avoid the possible overfitting at this stage, dropout layers are incorporated after each fully-connected layer. This fully-connected module is followed by a convolutional module with three 1D convolutional layers (\texttt{Conv1d} in PyTorch), each followed by a max-pooling layer and a ReLU activation, serving to extract the potential spatial features from each wavelength value. 

Subsequently, another fully-connected module (\texttt{rf\_fc}), consisting of fully connected layers and dropout layers mirroring the  \texttt{rec\_fc} module, is employed for the detailed reconstruction of finer spectral information. Furthermore, a residual connection links the initial output from \texttt{rec\_fc} with the detailed output from \texttt{rf\_fc}, to improve the quality of the final prediction without losing the key spectral features from the initial reconstruction. A sigmoid function is applied at the end to ensure the final output spectrum is smooth and continuous.

\section{Results}\label{section:experiments}

In this section, we illustrate the performance of our \textit{Sim2Real} approach on real-world data by comparing both the test accuracy and the inference time with a state-of-the-art optimization-based method: NNLS with Total Variation regularization (NNLS-TV)\cite{NNLS_TV}.

\subsection{\edit{Experimental Setup}} \label{subsec:exp_setup}
\edit{Our proposed DL-based model, \textit{ReSpecNN}, in our \textit{Sim2Real} setting was solely trained on the device-informed simulated data, generated according to \Cref{algorithm:data_aug}. To prepare the model for subsequent experiments, a total of $20,000$ simulated data pairs
($\bm{y}_{\textsf{aug}}$, $\bm{x}_{\textsf{sim}}$) were used.}
To address the potential scaling difference between simulated and real inputs encoded signal values, the \textit{log-min-max normalization} transformation is applied to convert the raw inputs $\bm{y}$ to $\widehat{\bm{y}}$ before fed into the neural network:

\begin{align}
    \widehat{\bm{y}} = \frac{\bm{z} - \min(\bm{z})}{\max(\bm{z}) - \min(\bm{z})},\quad \mbox{where} \quad 
    \bm{z} = \log\left(\bm{y}\right)
\end{align}\label{eq:log-min-max}

In the simulated data generation process, a batch size of $256$ is utilized. During the hierarchical data augmentation process, we set $S=2$ and $T=4$. For the noise control parameters, we chose $\alpha=5\times10^{-2}$ and $\sigma_{\epsilon} = 1\times10^{-5}$. \edit{In practice, both $S$ and $T$ could be determined empirically. For instance, we trained one model for each $S = {0, 1, 2, 3}$ until convergence and obtained corresponding MAEs of ${16.1, 1.96, 1.23, 1.94}$. Consequently, we selected $S = 2$. It's worth noting that as $S$ increases, the size of the training data also increases. Therefore, for more efficient training, our model favors smaller values of $S$. The noise level $\alpha$ and $\theta_{\epsilon}$ are determined similarly, either by sweeping between values or by employing a binary search method within an interval to find the appropriate value.} During the training stage, we employed the \texttt{MSELoss} function in \texttt{PyTorch} as our training loss. The optimizer used was \textit{Adam} with a learning rate of $3\times10^{-4}$. 


\begin{figure}[t]
    \centering 
    \includegraphics[width=\linewidth]{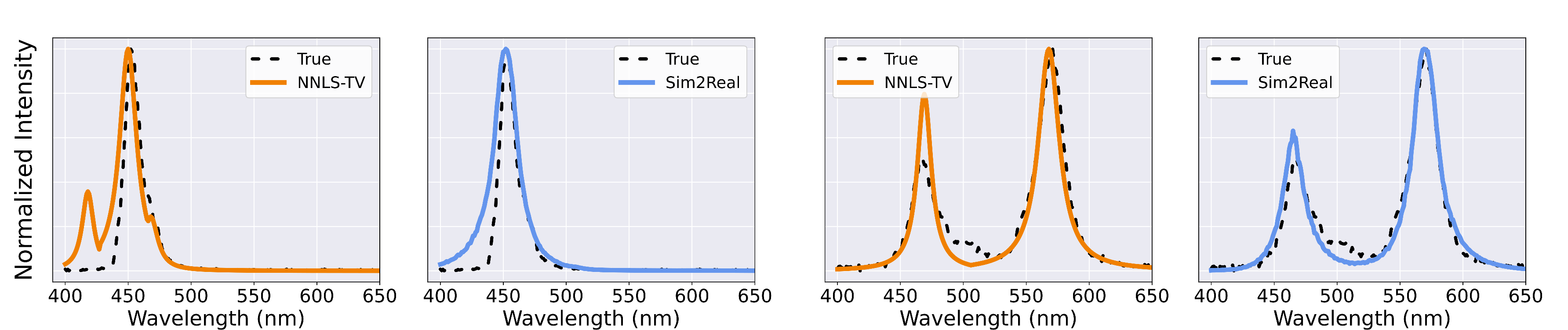}
    \caption{\textbf{Visualization of the reconstructed spectral signals using real-world data} (Left: a single-peak sample, Right: a double-peak sample) produced by NNLS-TV\cite{NNLS_TV} and our Sim2Real approach. The corresponding ground-truth spectra are measured by a commercial spectrometer. Our Sim2Real approach achieves a comparable performance with NNLS-TV. \edit{For any results with asymmetric peaks, the peak position is defined as the wavelength at which the maximum intensity occurs.}} 
    \label{fig:samples}
\end{figure}


\subsection{Real-world Data Evaluation}

\edit{To demonstrate the performance of our \textit{Sim2Real} framework, we conducted evaluations using a real-world dataset, which is collected with a portable spectrometer device\cite{NNLS_TV}. This real-world dataset includes nine single-peak and five multiple-peak spectral signal samples, covering the wavelength range from $400$ nm to $650$ nm needed for the spectrometer's targeted mobile application. The diverse peak profiles and the wide wavelength span ensure that the dataset is well-suited for testing the model under realistic conditions.}

To evaluate the effectiveness of our proposed \textit{Sim2Real} framework in comparison with the optimization-based method NNLS-TV\cite{NNLS_TV}, we focused mainly on two critical aspects: the accuracy of the spectral peak location and relative peak intensity. 
\edit{Instead of attempting to predict the entire spectral shape which would require a large number of deployed spectral encoders
\footnote{\edit{In fact, the number of spectral encoders is the most important factor determining the spectral resolution, which in turn dictates the criterion of the Relative Peak Position Error.}}
, 
our goal is to efficiently predict the most useful parameters for the mobile application, the spectral peak locations, using the least number of spectral encoders.}
\edit{Thus, we chose the following Relative Peak Position Error as our evaluation criterion, defined as} 
\begin{align*}
    \text{Rel. error} \ \Delta \lambda / \lambda := (\lambda_{\textsf{reconstructed}} - \lambda) / \lambda,
\end{align*}
for each reconstruction spectral signal. Here, $\lambda_{\textsf{reconstructed}}$ denotes the wavelength position of the peak for the reconstructed spectral signal, while $\lambda$ corresponds to the position of the ground truth. 


Specific to multi-peaks spectral data, we also measured the Relative Peak Intensity Error defined as 
\begin{align*}
    \text{Rel. error} \ \Delta I / I := (I_{\textsf{reconstructed}} - I) / I,
\end{align*}
where $I_{\textsf{reconstructed}}$ denotes the normalized peak intensity of the reconstructed spectral signal, while $I$ corresponds to the ground truth.


\Cref{fig:samples} presents two spectrum reconstruction examples using our proposed deep neural network and the NNLS-TV\cite{NNLS_TV} approach. When compared with the ground truths obtained by a commercial spectrometer, our approach exhibits a reliable performance on our miniature chip-scale spectrometer.

\begin{figure}[t]
    \centering
    \includegraphics[width=0.5\linewidth]{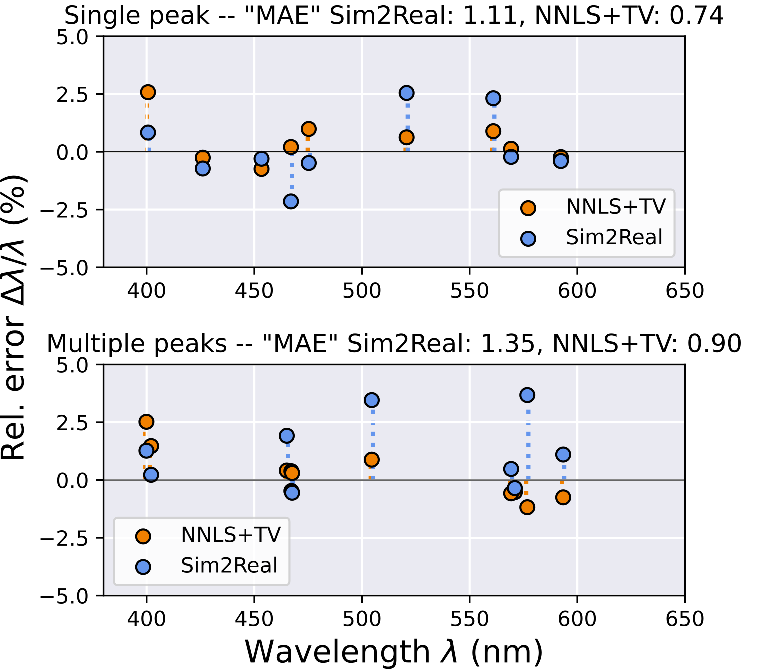}
    \caption{\textbf{The relative peak position errors} (vs. wavelength in nm) for single-peak real data (Top) and multiple-peak real data (Bottom).
    The plot title shows the \underline{m}ean \underline{a}bsolute \underline{e}rror (Abbrev.\ ``MAE'') for each approach. 
    }
    \label{fig:relative-peak-positions}
\end{figure}

For peak position predictions, \Cref{fig:relative-peak-positions} show that our approach yields a result comparable with NNLS-TV\cite{NNLS_TV} \edit{on aforementioned real-world dataset\cite{NNLS_TV}.}
\edit{In \Cref{fig:relative-peak-positions}, we calculated the Mean Absolute Error (MAE), defined as the average of the absolute values of the Relative Peak Position Error across all the spectral data samples.}
While offering a significantly faster inference speed (details to be discussed in the next section), our model can still maintain the relative errors within -5 to 5 percent. 

\begin{figure}[t]
    \centering 
    \includegraphics[width=0.5\linewidth]{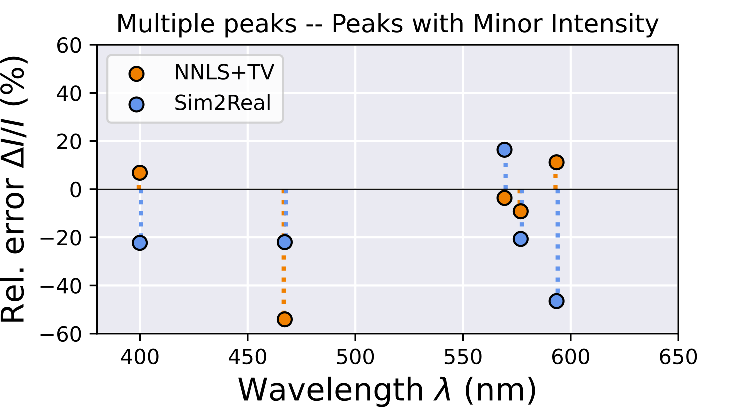}
    
    \caption{\textbf{The relative  intensity errors} (vs. wavelength in nm) for the subset of ``minor'' peaks with relatively lower intensities.}
    \label{fig:ratio}
\end{figure}

Regarding the results for relative peak intensity, which represent a more difficult challenge as illustrated in \Cref{fig:ratio}, our method effectively constrained the maximum relative error in intensity to under 50 percent. In contrast, the NNLS-TV approach occasionally exhibits significant prediction errors on certain spectral data samples.

\subsection{Execution Time Analysis for Reconstruction Methods}


Our miniature chip-scale spectrometers have been integrated into wearable health monitoring devices. In such scenarios, the inference time for reconstruction often emerges as a critical factor for achieving fast real-time health monitoring. Additionally, our pre-trained Sim2Real method offers the potential to conserve battery life, as it requires only one forward pass for inference, unlike NNLS-TV which requires sophisticated software or smart programs to iteratively solve it. Overall, prioritizing fast inference time is advantageous for real-time monitoring and extended battery life because of the low computational/energy costs involved in inference.

To demonstrate our superior inference time, we compared the execution time of our pre-trained model with the NNLS-TV solver using the real dataset\cite{NNLS_TV}. The results in \Cref{tab:time_compare} show that our model significantly reduces inference time. \edit{When conducting the execution time experiments in \Cref{tab:time_compare}, we ensured fairness by utilizing only the CPU for both the NNLS-TV and our DL approach. Moreover, tests were performed on identical hardware setups to guarantee that the comparison solely reflects the computational efficiency of each method.}


\begin{table}[hbt]
\centering
\begin{tabular}{ |c|c|c|c| } 
\hline
& \makecell{NNLS+TV\\(MATLAB)} & \makecell{NNLS+TV\\(SciPy)}  &  \makecell{\textbf{Ours}\\\textbf{(PyTorch)}} \\
\hline
\makecell{\textbf{Exec.}\\\textbf{Time (ms)}} & $ 10.634 \pm 0.126 $  & $ 7.965 \pm 0.230$ & $\bm{1.134 \pm 0.330}$ \\
\hline
\end{tabular}
\caption{\textbf{The average execution time} (per sample, in milliseconds) for the NNLS-TV\cite{NNLS_TV} method and our proposed approach on the real dataset\cite{NNLS_TV}. The NNLS-TV method\cite{NNLS_TV} was originally implemented with the MATLAB solver \texttt{lsqnonneg} in the paper, and to ensure a fair comparison with our model implemented in PyTorch, we also implemented their method using \texttt{scipy.optimize.nnls} in SciPy.}
\label{tab:time_compare}
\end{table}

\subsection{Domain Shift in Reconstructive Spectroscopy} \label{subsec:diss_domain_gap}


As previously discussed in \Cref{sec:domain-gap}, the domain shift usually exists between the device-informed simulated data $\bm{y}_{\textsf{sim}}$ and the real-world data $\bm{y}_{\textsf{real}}$. Many works only focus on improving the reconstruction accuracy within the datasets of the same distribution (i.e. training and testing exclusively on either simulated data or specific real-measured datasets) without attempting to bridge the domain gap between them.
For instance, \textit{ResCNN}\cite{kim2022compressive} has been demonstrated to achieve excellent spectral reconstruction performance and maintain stable results even under some noisy conditions. It preprocessed the input data from $\bm{y}$ to $\widetilde{\bm{x}} := \bm{R}^\dagger \bm{y}$ ($\bm{R}^\dagger$ is the pseudoinverse of $\bm{R}$) for improved results. However, \textit{ResCNN} trained only on simulated data does not perform optimally when directly applied to real-world data\cite{NNLS_TV}.




\Cref{fig:comparison-with-kim2022compressive} illustrates a comparison of the root mean square error (RMSE) under the \textit{Sim2Real} setting between the spectral signals reconstructed by \textit{ResCNN} and our \textit{ReSpecNN}, against the ground truths for both the device-informed simulated data and the real data collected from our spectrometer\cite{NNLS_TV}. While \textit{ResCNN} nearly perfectly fits the simulated data (with a lower RMSE), its performance on the real dataset\cite{NNLS_TV} tends to drop significantly under the \textit{Sim2Real} training setting, which confirms the domain shift phenomenon.

\section{Conclusion and Discussion}
In this paper, we introduce a novel \textit{Sim2Real} framework for spectral signal reconstruction in spectroscopy, focusing on the sampling efficiency and inference time.
Throughout the training process, only a single measurement of the response matrix $\bm{R}$ for the spectrometer device is required, with all other training data being simulated and generated. To address the domain shift between the real-world data and the device-informed simulated data, our \textit{Sim2Real} framework introduces the hierarchical data augmentation approach to train our deep learning model. 
Furthermore, our neural network, \textit{ReSpecNN}, which is trained exclusively on such augmented simulated data, is specifically designed for the reconstruction of real spectral signals.
In our experiments, even with the simplest Gaussian noise augmentation, our \textit{Sim2Real} method has achieved a 10-fold speed-up during the inference compared to the state-of-the-art optimization-based solver NNLS-TV\cite{NNLS_TV}, while demonstrating on par performance in terms of the solution quality. \edit{Although using only Gaussian noises for augmentation presents a limitation, the flexibility of our data augmentation suggests that improvements on the perturbation model of the response matrix $\bm R$ may further improve our \textit{Sim2Real} framework by experimentally determining the noise distribution.} 

\edit{In short, our hierarchical data augmentation strategy significantly improves the model robustness on spectral signal reconstruction in spectroscopy by realistically simulating the noise that can occur within the spectrometer device. However, it does have limitations. For example, in cases of extreme outliers that arise in some of the real spectral data, even after extensive data augmentation (corresponding to the high noise level in our method), those spectral data may still not be represented adequately. This leads to suboptimal performance in spectrum reconstruction for these extreme cases. Nevertheless, this limitation is a common challenge across data augmentation techniques. 
}

\edit{Moving forward, to better handle outliers and further improve the robustness and accuracy of our model, we plan to apply the following approaches.} First, it would be interesting to explore some improved noise patterns within our data augmentation process, for instance, incorporating the idea of adversarial training of deep networks\cite{goodfellow2014explaining}. Furthermore, while we focus on scenarios without labeled real training data, limited labeled data may be available in practice. These data can fine-tune our model, but the small size risks overfitting. To mitigate this while employing these data, we can use simulated training data for regularization during fine-tuning, as suggested in the recent study \cite{ruiz2023dreambooth}.


\edit{Regarding the model scalability, the input size in the spectral reconstruction problem depends on the number of spectral encoders used within the spectrometer device, which varies based on specific problem setup and device design. In this paper, the number of spectral encoders is fixed at 16. Looking ahead, we could still explore our model scalability under potentially different problem setups, for instance, the reconstruction with polarized spectral encoded signals.}

\section*{Acknowledgment}
JYC, PYL, QQ, and YW acknowledge support from NSF CAREER CCF-2143904, NSF CCF-2212066, MIDAS PODS grant, and a gift grant from KLA. PCK acknowledges the support from NSF ECCS-2317047 for device fabrication and spectral data collection. YW also acknowledges support from the Eric and Wendy Schmidt AI in Science Postdoctoral Fellowship, a Schmidt Futures program.

\section*{Data Availability Statement}
The testing spectral data and our experimental results are available at \url{https://github.com/j1goblue/Rec_Spectrometer}.

\pagebreak

\begin{thebibliography}{10}

\bibitem{chang2008estimation}
Cheng-Chun Chang and Heung-No Lee.
\newblock On the estimation of target spectrum for filter-array based spectrometers.
\newblock {\em Optics Express}, 16(2):1056--1061, 2008.

\bibitem{Kurokawa2011}
Umpei Kurokawa, Byung~Il Choi, and Cheng~Chun Chang.
\newblock Filter-based miniature spectrometers: Spectrum reconstruction using adaptive regularization.
\newblock {\em IEEE Sensors Journal}, 11:1556--1563, 2011.

\bibitem{bao2015colloidal}
Jie Bao and Moungi~G Bawendi.
\newblock A colloidal quantum dot spectrometer.
\newblock {\em Nature}, 523(7558):67--70, 2015.

\bibitem{Wang2014e}
Zhu Wang and Zongfu Yu.
\newblock Spectral analysis based on compressive sensing in nanophotonic structures.
\newblock {\em Optics Express}, 22:25608, 10 2014.

\bibitem{Wang2019}
Zhu Wang, Soongyu Yi, Ang Chen, Ming Zhou, Ting~Shan Luk, Anthony James, John Nogan, Willard Ross, Graham Joe, Alireza Shahsafi, Ken~Xingze Wang, Mikhail~A. Kats, and Zongfu Yu.
\newblock Single-shot on-chip spectral sensors based on photonic crystal slabs.
\newblock {\em Nature Communications}, 10:1020, 12 2019.

\bibitem{rs2}
Cheolsun Kim, Woong-Bi Lee, Soo~Kyung Lee, Yong~Tak Lee, and Heung-No Lee.
\newblock Fabrication of 2d thin-film filter-array for compressive sensing spectroscopy.
\newblock {\em Optics and Lasers in Engineering}, 115:53--58, 2019.

\bibitem{Sarwar2020}
Tuba Sarwar, Srinivasa Cheekati, Kunook Chung, and Pei-Cheng Ku.
\newblock On-chip optical spectrometer based on gan wavelength-selective nanostructural absorbers.
\newblock {\em Applied Physics Letters}, 116:081103, 2 2020.

\bibitem{UCLA_paper}
Calvin Brown, Artem Goncharov, Zachary~S. Ballard, Mason Fordham, Ashley Clemens, Yunzhe Qiu, Yair Rivenson, and Aydogan Ozcan.
\newblock Neural network-based on-chip spectroscopy using a scalable plasmonic encoder.
\newblock {\em ACS Nano}, 15(4):6305--6315, 2021.
\newblock PMID: 33543919.

\bibitem{chip_rs1}
Jiajun Meng, Jasper~J. Cadusch, and Kenneth~B. Crozier.
\newblock Plasmonic mid-infrared filter array-detector array chemical classifier based on machine learning.
\newblock {\em ACS Photonics}, 8(2):648--657, 2021.

\bibitem{chip_rs2}
Shang Zhang, Yuhan Dong, Hongyan Fu, Shao-Lun Huang, and Lin Zhang.
\newblock A spectral reconstruction algorithm of miniature spectrometer based on sparse optimization and dictionary learning.
\newblock {\em Sensors}, 18(2):644, Feb 2018.

\bibitem{candes2006stable}
Emmanuel~J Candes, Justin~K Romberg, and Terence Tao.
\newblock Stable signal recovery from incomplete and inaccurate measurements.
\newblock {\em Communications on Pure and Applied Mathematics: A Journal Issued by the Courant Institute of Mathematical Sciences}, 59(8):1207--1223, 2006.

\bibitem{RUDIN1992259}
Leonid~I. Rudin, Stanley Osher, and Emad Fatemi.
\newblock Nonlinear total variation based noise removal algorithms.
\newblock {\em Physica D: Nonlinear Phenomena}, 60(1):259--268, 1992.

\bibitem{NNLS_TV}
Tuba Sarwar, Can Yaras, Xiang Li, Qing Qu, and Pei-Cheng Ku.
\newblock Miniaturizing a chip-scale spectrometer using local strain engineering and total-variation regularized reconstruction.
\newblock {\em Nano Letters}, 22(20):8174--8180, 2022.
\newblock PMID: 36223431.

\bibitem{Li23}
Pengyu Li, Can Yaras, Tuba Sarwar, Pei-Cheng Ku, and Qing Qu.
\newblock Accelerating deep learning in reconstructive spectroscopy using synthetic data.
\newblock In {\em CLEO 2023}, page JTu2A.71. Optica Publishing Group, 2023.

\bibitem{kim2022compressive}
Cheolsun Kim, Dongju Park, and Heung-No Lee.
\newblock Compressive sensing spectroscopy using a residual convolutional neural network.
\newblock {\em Sensors}, 20(3):594, Jan 2020.

\bibitem{zhang2020denoising}
Jinhui Zhang, Xueyu Zhu, and Jie Bao.
\newblock Denoising autoencoder aided spectrum reconstruction for colloidal quantum dot spectrometers.
\newblock {\em IEEE sensors journal}, 21(5):6450--6458, 2020.

\bibitem{arora2019implicit}
Sanjeev Arora, Nadav Cohen, Wei Hu, and Yuping Luo.
\newblock Implicit regularization in deep matrix factorization.
\newblock {\em Advances in Neural Information Processing Systems}, 32, 2019.

\bibitem{pan2009survey}
Sinno~Jialin Pan and Qiang Yang.
\newblock A survey on transfer learning.
\newblock {\em IEEE Transactions on knowledge and data engineering}, 22(10):1345--1359, 2009.

\bibitem{tzeng2017adversarial}
Eric Tzeng, Judy Hoffman, Kate Saenko, and Trevor Darrell.
\newblock Adversarial discriminative domain adaptation.
\newblock In {\em Proceedings of the IEEE conference on computer vision and pattern recognition}, pages 7167--7176, 2017.

\bibitem{ganin2015unsupervised}
Yaroslav Ganin and Victor Lempitsky.
\newblock Unsupervised domain adaptation by backpropagation.
\newblock In {\em International conference on machine learning}, pages 1180--1189. PMLR, 2015.

\bibitem{zhang2018collaborative}
Weichen Zhang, Wanli Ouyang, Wen Li, and Dong Xu.
\newblock Collaborative and adversarial network for unsupervised domain adaptation.
\newblock In {\em Proceedings of the IEEE conference on computer vision and pattern recognition}, pages 3801--3809, 2018.

\bibitem{finn2017model}
Chelsea Finn, Pieter Abbeel, and Sergey Levine.
\newblock Model-agnostic meta-learning for fast adaptation of deep networks.
\newblock In {\em International conference on machine learning}, pages 1126--1135. PMLR, 2017.

\bibitem{vinyals2016matching}
Oriol Vinyals, Charles Blundell, Timothy Lillicrap, Daan Wierstra, et~al.
\newblock Matching networks for one shot learning.
\newblock {\em Advances in neural information processing systems}, 29, 2016.

\bibitem{snell2017prototypical}
Jake Snell, Kevin Swersky, and Richard Zemel.
\newblock Prototypical networks for few-shot learning.
\newblock {\em Advances in neural information processing systems}, 30, 2017.

\bibitem{girshick2014rich}
Ross Girshick, Jeff Donahue, Trevor Darrell, and Jitendra Malik.
\newblock Rich feature hierarchies for accurate object detection and semantic segmentation.
\newblock In {\em Proceedings of the IEEE conference on computer vision and pattern recognition}, pages 580--587, 2014.

\bibitem{zoph2018learning}
Barret Zoph, Vijay Vasudevan, Jonathon Shlens, and Quoc~V Le.
\newblock Learning transferable architectures for scalable image recognition.
\newblock In {\em Proceedings of the IEEE conference on computer vision and pattern recognition}, pages 8697--8710, 2018.

\bibitem{dai2015semi}
Andrew~M Dai and Quoc~V Le.
\newblock Semi-supervised sequence learning.
\newblock {\em Advances in neural information processing systems}, 28, 2015.

\bibitem{howard2018universal}
Jeremy Howard and Sebastian Ruder.
\newblock Universal language model fine-tuning for text classification.
\newblock In Iryna Gurevych and Yusuke Miyao, editors, {\em Proceedings of the 56th Annual Meeting of the Association for Computational Linguistics (Volume 1: Long Papers)}, pages 328--339, Melbourne, Australia, July 2018. Association for Computational Linguistics.

\bibitem{devlin2018bert}
Jacob Devlin, Ming-Wei Chang, Kenton Lee, and Kristina Toutanova.
\newblock {BERT}: Pre-training of deep bidirectional transformers for language understanding.
\newblock In Jill Burstein, Christy Doran, and Thamar Solorio, editors, {\em Proceedings of the 2019 Conference of the North {A}merican Chapter of the Association for Computational Linguistics: Human Language Technologies, Volume 1 (Long and Short Papers)}, pages 4171--4186, Minneapolis, Minnesota, June 2019. Association for Computational Linguistics.

\bibitem{chen2021representation}
Xinyang Chen, Sinan Wang, Jianmin Wang, and Mingsheng Long.
\newblock Representation subspace distance for domain adaptation regression.
\newblock In {\em ICML}, pages 1749--1759, 2021.

\bibitem{nejjar2023dare}
Ismail Nejjar, Qin Wang, and Olga Fink.
\newblock Dare-gram: Unsupervised domain adaptation regression by aligning inverse gram matrices.
\newblock In {\em Proceedings of the IEEE/CVF Conference on Computer Vision and Pattern Recognition}, pages 11744--11754, 2023.

\bibitem{dong2015image}
Chao Dong, Chen~Change Loy, Kaiming He, and Xiaoou Tang.
\newblock Image super-resolution using deep convolutional networks.
\newblock {\em IEEE transactions on pattern analysis and machine intelligence}, 38(2):295--307, 2015.

\bibitem{kulkarni2016reconnet}
Kuldeep Kulkarni, Suhas Lohit, Pavan Turaga, Ronan Kerviche, and Amit Ashok.
\newblock Reconnet: Non-iterative reconstruction of images from compressively sensed measurements.
\newblock In {\em Proceedings of the IEEE conference on computer vision and pattern recognition}, pages 449--458, 2016.

\bibitem{ongie2020deep}
Gregory Ongie, Ajil Jalal, Christopher~A Metzler, Richard~G Baraniuk, Alexandros~G Dimakis, and Rebecca Willett.
\newblock Deep learning techniques for inverse problems in imaging.
\newblock {\em IEEE Journal on Selected Areas in Information Theory}, 1(1):39--56, 2020.

\bibitem{goodfellow2014explaining}
Ian~J. Goodfellow, Jonathon Shlens, and Christian Szegedy.
\newblock Explaining and harnessing adversarial examples.
\newblock {\em International Conference on Learning Representations (ICLR)}, 2015.

\bibitem{ruiz2023dreambooth}
Nataniel Ruiz, Yuanzhen Li, Varun Jampani, Yael Pritch, Michael Rubinstein, and Kfir Aberman.
\newblock Dreambooth: Fine tuning text-to-image diffusion models for subject-driven generation.
\newblock In {\em Proceedings of the IEEE/CVF Conference on Computer Vision and Pattern Recognition}, pages 22500--22510, 2023.

\end{thebibliography}

\end{document}